\documentclass[runningheads]{llncs}
\usepackage[T1]{fontenc}
\usepackage{graphicx}
\usepackage{algorithm}%
\usepackage{algorithmicx}%
\usepackage{algpseudocode}
\usepackage{url}
\usepackage{amsmath} 
\begin{document}
\title{SentiDrop: A Multi-Modal Machine Learning model for Predicting Dropout in Distance Learning}
\titlerunning{Abbreviated paper title}
%

\author{Meriem Zerkouk\inst{1} \and
Miloud Mihoubi\inst{1} \and
Belkacem Chikhaoui\inst{1} 
}
%
%
\institute{
Artificial Intelligence Institute, University of Téluq, 5800, rue Saint-Denis, Montreal, Quebec, H2S 3L5, Canada\\
}
\maketitle              
\begin{abstract}
School dropout is a serious problem in distance learning, where early detection is crucial for effective intervention and student perseverance. Predicting student dropout using available educational data is a widely researched topic in learning analytics. Our partner’s distance learning platform highlights the importance of integrating diverse data sources, including socio-demographic data, behavioral data, and sentiment analysis, to accurately predict dropout risks. In this paper, we introduce a novel model that combines sentiment analysis of student comments using the Bidirectional Encoder Representations from Transformers (BERT) model with socio-demographic and behavioral data analyzed through Extreme Gradient Boosting (XGBoost). We fine-tuned BERT on student comments to capture nuanced sentiments, which were then merged with key features selected using feature importance techniques in XGBoost. Our model was tested on unseen data from the next academic year, achieving an accuracy of 84\%, compared to 82\% for the baseline model. Additionally, the model demonstrated superior performance in other metrics, such as precision and F1-score. The proposed method could be a vital tool in developing personalized strategies to reduce dropout rates and encourage student perseverance.

\end{abstract}
\section{Introduction}

The rapid growth of distance learning has significantly improved access to education and redefined the educational experience by offering flexibility and opportunities that traditional classroom environments often cannot provide. However, this flexibility also introduces challenges, particularly in addressing the critical issue of student dropout, which has far-reaching implications for students, educational institutions, and society.   Student dropout is a global problem with significant negative consequences for individuals, families, and communities. For instance, in Quebec, Canada, approximately 10,000 young people leave school without obtaining a first diploma each year, resulting in a lifetime cost of \$120 million per annual cohort to society \cite{LeDevoir2023}. The economic impact is profound, as school dropouts face significant difficulties finding employment and typically earn lower incomes throughout their lives. This issue is not only an economic concern but also has social, academic, political, and financial ramifications, making it imperative to develop effective strategies to prevent dropouts and support students in their educational pursuits. The COVID-19 pandemic has further accelerated the adoption of online platforms, underscoring the importance of technology-based solutions in maintaining educational access during crises. Despite advances in early detection methods \cite{Berens2018}, which often focus on school absenteeism and poor academic performance as critical indicators of dropout risk, other factors also contribute significantly to school dropout. Internal factors such as learning difficulties and mental or behavioral problems, and external factors like financial difficulties, housing instability, or insufficient family support, play substantial roles. This complexity highlights the need for more advanced, data-driven approaches to capture a broader range of factors influencing dropout risks. Accurately predicting student dropout is especially critical in online education, where early interventions can significantly improve student retention. ML has emerged as a powerful tool for this purpose, offering the potential for early warnings by analyzing large datasets to identify patterns and predictors of dropout. However, traditional ML models often face challenges, including class imbalance and the "black box" nature, which makes it difficult to interpret results and guide interventions effectively. Our research addresses these challenges by introducing a novel predictive model that uniquely combines sentiment analysis, socio-demographic data, and behavioral data, forming a more holistic and nuanced approach to dropout prediction. Our model captures objective academic and demographic indicators of dropout risk and integrates subjective elements related to students' emotional experiences within the learning environment. We assess the sentiments expressed in students' online interactions by employing advanced sentiment analysis techniques, ensuring a more granular analysis. We have implemented robust cross-validation techniques to enhance the model's reliability and prevent overfitting, especially when students have multiple comments. These techniques ensure that comments from the same student are not split across training and test sets, thereby preserving the integrity of our model’s performance assessment and its generalizability to new, unseen data.

The main research questions of this study are:

\begin{enumerate}
    \item How can we develop ML models and algorithms that are effective in predicting student dropout in distance learning?
    \item Which features are most effective for predicting student dropout with high accuracy and performance?
\end{enumerate}

This study presents three main contributions:

\begin{enumerate}
    \item \textbf{Integration of Diverse Data Sources}: Combining socio-demographic, behavioral, and sentiment data improves the accuracy of student dropout predictions by providing a more comprehensive understanding of the factors contributing to dropout risks. The model offers a more nuanced and accurate prediction by capturing objective indicators (academic performance and demographics) and subjective elements (emotional well-being).
    
    \item \textbf{Practical Implications of Sentiment Analysis}: Using sentiment analysis in educational platforms to predict dropout risks enables the monitoring of students' emotional well-being in real time, allowing for early and targeted interventions. Sentiment analysis can reveal underlying issues that traditional academic metrics might miss, thereby improving student retention strategies.
    
    \item \textbf{Enhanced Predictive Performance}: The proposed model outperforms existing predictive models by integrating diverse data sources, which enhances its accuracy and robustness. In practical application, it provides more actionable insights, allowing educators to implement more effective dropout prevention measures. The model's use of sentiment analysis, in particular, offers a novel dimension that existing models typically lack, leading to better identification of at-risk students.
\end{enumerate}

\section{Related work}

Machine learning (ML) has emerged as a powerful technique for predicting student dropout in distance learning platforms \cite{Pek2023}. ML models analyze a wide range of features, including academic performance, behavioral patterns, and sociodemographic data, to accurately identify at-risk students. Careful selection of relevant features enhances these models, enabling more precise predictions and providing deeper insights into the causes of student dropout. Furthermore, incorporating emotional and experiential data, such as sentiments expressed in student comments, strengthens these models, allowing for more personalized and effective interventions. The potential of ML to revolutionize education should inspire and motivate educators and researchers to develop advanced predictive tools.

As noted by Karabacak et al. \cite{Karabacak2023} \cite{Karabacak2024}, various ML approaches, including classification and regression methods, neural networks, Bayesian networks, support vector machines, and logistic regression, have been explored to tackle the challenge of predicting student performance. 

Kemper et al. \cite{Kemper2020} demonstrated the effectiveness of these approaches in predicting student dropout in higher education, with the random forest model outperforming other models such as decision trees and neural networks.

 Solís et al. \cite{Solis2018} presented an ML model designed to predict student dropout from university courses by integrating both academic and demographic factors as key predictors. This body of work underscores the importance of selecting appropriate ML techniques and feature sets to enhance the accuracy and effectiveness of dropout prediction models, ultimately contributing to improved student retention and success.  

Sani et al. \cite{Sani2020} demonstrated the effectiveness of the Random Forest model in accurately predicting student dropout among Malaysia's low-income B40 group, offering a powerful tool for targeted interventions in higher education.

Niyogisubizo et al. \cite{Niyogisubizo2022} address student dropout using a novel stacking ensemble model that combines Random Forest, XGBoost, Gradient Boosting, and Feed-forward Neural Networks. This ensemble model outperforms individual models in predicting student dropout by utilizing a unique dataset from Constantine the Philosopher University in Nitra.

In summary, our research significantly advances the field of dropout prediction by introducing a novel approach that integrates sentiment analysis with sociodemographic and behavioral data. This multimodal strategy offers a fresh perspective on understanding, predicting, and encouraging student perseverance in online education. The unique application of the BERT algorithm for sentiment classification and XGBoost for predictive analysis demonstrates an innovative use of these advanced tools in the educational domain. Our comparative analyses highlight the specific improvements and enhancements brought by this method, underscoring its potential to contribute significantly to early dropout prediction, support student perseverance, and develop personalized interventions. Consequently, our model is likely to lead to a substantial reduction in student dropout rates and a marked improvement in student success within higher education platforms.

\section{Methodology}

Our approach involves a multi-module system to predict student dropout by leveraging socio-demographic, behavioral, and sentiment data. The methodology consists of three interconnected modules: Data Pre-processing, Data Merging, and Prediction Model Development.

\subsubsection{Data Pre-processing}

The initial phase of our methodology focuses on data pre-processing, which is crucial for ensuring the quality and reliability of the input data. This step involves:

\paragraph{Data Cleaning:} We systematically eliminate inconsistencies, handle missing values, and correct errors within the raw dataset. This ensures that the data is robust and ready for further analysis.

Mathematically, let $\mathbf{x}_i$ denote the feature vector for the $i$-th student, where $\mathbf{x}_i = [x_{i1}, x_{i2}, \dots, x_{im}]$ represents $m$ features associated with that student.

\paragraph{Feature Normalization:} Depending on the requirements of the subsequent models, feature normalization or scaling is applied to ensure that all features contribute equally to the model's predictions.

\subsubsection{Data Merging}

This module integrates two critical components to enhance the predictive power of our model:

\paragraph{Dropout Detection Model:} This model utilizes socio-demographic, behavioral, and sentiment data to identify students at risk of dropping out. It leverages our previous research, which demonstrated the effectiveness of combining these data types for dropout prediction.

\paragraph{Sentiment Analysis Model:} Sentiment analysis modele consists to converting qualitative student feedback into quantitative features, capturing emotional and experiential nuances that are often missed by purely behavioral or sociodemographic data ~\cite{Khanam2023}~\cite{Baragash2021}. This component classifies student sentiments into positive, negative, or neutral categories based on their interactions within the learning platform. The sentiment data provides valuable insights into students' emotional states, which can be strong indicators of their likelihood to drop out.
\begin{algorithm}
\caption{Sentiment Analysis for Dropout Prediction}
\begin{algorithmic}[1]
\For{each student comment at time $t$}
    \State Compute sentiment score $S_i(t)$ using BERT
    \State Aggregate scores monthly to calculate average sentiment $\bar{S}(M)$
\EndFor
\State Determine the number of comments per month $N_M$
\State Calculate monthly average sentiment:
\State $\bar{S}(M) = \frac{1}{N_M} \sum_{i=1}^{N_M} S_i(M)$
\State Use paired t-test to compare sentiment:
\State $t = \frac{\bar{X}_D - \mu_0}{s_D / \sqrt{n}}$
\State Integrate sentiment data with primary dataset
\State Identify at-risk students and suggest interventions
\end{algorithmic}
\end{algorithm}

The Algorithm 1 involves computing sentiment scores for each student comment using the BERT algorithm and aggregating these scores monthly.

\subsubsection{Prediction Model Development}

The final module in our methodology is the development of a predictive model that synthesizes insights from both the dropout detection and sentiment analysis model:

\paragraph{Dataset Representation:} The dataset $D$ contains $n$ instances, where each instance corresponds to a student. Let $D = \{(x_i, y_i)\}_{i=1}^n$, where $x_i$ represents the feature vector of the $i$-th student, and $y_i$ denotes the binary dropout status (1 for dropout, 0 for active).

\subsubsection{Feature Importance and Selection}

 It  improve the interpretability and performance of our predictive model, we utilize SHapley Additive exPlanations (SHAP) to calculate the importance of each feature in the dataset. SHAP values provide a unified measure of feature importance, derived from cooperative game theory, which explains the contribution of each feature to the prediction.

Let $D''$ be the preprocessed dataset, and $\mathbf{x}_i = [x_{i1}, x_{i2}, \dots, x_{in}]$ be the feature vector for the $i$-th student, where $n$ is the total number of features. The SHAP value for the $j$-th feature of the $i$-th student is denoted as $\phi_{ij}$. The SHAP values are computed as:

\begin{equation}
\phi_{ij} = \sum_{S \subseteq F \setminus \{j\}} \frac{|S|!(m - |S| - 1)!}{m!} \left[ f_S(x_i) - f_{S \cup \{j\}}(x_i) \right]
\end{equation}

where $f_S(x_i)$ represents the model trained on a subset $S$ of features, and $f_{S \cup \{j\}}(x_i)$ represents the model with the $j$-th feature added to $S$. The SHAP value $\phi_{ij}$ captures the contribution of feature $x_{ij}$ to the prediction for student $i$.

Once the SHAP values are computed, we rank the features based on their average absolute SHAP values across all instances:

\begin{equation}
\text{Importance}(j) = \frac{1}{n} \sum_{i=1}^{n} |\phi_{ij}|
\end{equation}

We then select the top $k$ features with the highest SHAP importance scores for the final model, ensuring that the most influential features are included in the prediction model.

\subsubsection{Model Training }

In developing our predictive model, we adopt an ensemble approach that combines the outputs of multiple machine-learning models to improve accuracy and robustness. Specifically, we train three distinct models: eXtreme Gradient Boosting (XGBoost), Random Forest (RF), and Logistic Regression (LR). Each model captures different aspects of the data, contributing to a more comprehensive prediction when combined.

For a given student $i$, let $P_{XG}(x_i)$, $P_{RF}(x_i)$, and $P_{LR}(x_i)$ represent the predicted dropout probabilities from the XGBoost, Random Forest, and Logistic Regression models, respectively. These probabilities are computed as:
\begin{equation}
P_{XG}(x_i) = \text{XGModel}(x_i), \quad P_{RF}(x_i) = \text{RFModel}(x_i), \quad P_{LR}(x_i) = \text{LRModel}(x_i)
\end{equation}

The final prediction for the dropout probability, $P_i$, is obtained by averaging the outputs from these three models:

\begin{equation}
P_i = \frac{P_{XG}(x_i) + P_{RF}(x_i) + P_{LR}(x_i)}{3}
\end{equation}

This ensemble approach mitigates the biases and weaknesses of individual models, leading to a more reliable and accurate prediction.

To evaluate the performance of the ensemble model, we define a loss function that measures the discrepancy between the predicted dropout probability $P_i$ and the actual dropout status $y_i$. A common choice for binary classification problems is the binary cross-entropy loss, which is defined as:

\begin{equation}
L(y_i, P_i) = -\left[ y_i \log(P_i) + (1 - y_i) \log(1 - P_i) \right]
\end{equation}

The objective is to minimize the total loss over all students in the dataset:

\begin{equation}
\text{minimize} \sum_{i=1}^{n} L(y_i, P_i)
\end{equation}

This optimization ensures the predictive model is fine-tuned to accurately reflect dropout risks among different students, enhancing its generalizability and effectiveness.

\paragraph{Objective:} Our goal is to develop a predictive function $f(x)$ that accurately predicts the dropout status $y_i$ given the feature vector $x_i$. This function merges outputs from the dropout detection ($DD(x_i)$) and sentiment analysis ($SA(x_i)$) models:

\begin{equation}
f(x_i) = \text{Predict}(x_i) = \text{Merge}(DD(x_i), SA(x_i))
\end{equation}

The $\text{Merge}(\cdot)$ function can be implemented using various ML techniques, such as logistic regression, decision trees, neural networks, or ensemble methods.

\paragraph{Loss Function and Optimization:} To quantify the performance of the model, we define a loss function $L(y_i, f(x_i))$, which measures the discrepancy between the predicted dropout probability $f(x_i)$ and the actual dropout status $y_i$.

The goal is to minimize the overall loss across all $n$ instances in the dataset:

\begin{equation}
\text{minimize} \sum_{i=1}^{n} L(y_i, f(x_i))
\end{equation}

This optimization is crucial for refining the model's predictions and ensuring that it generalizes well to unseen data.

\section{Experimental Results}

\subsection{Dataset description}
The dataset for this study was drawn from our partner’s distance learning platform, active between 2016 and 2023, and comprises 35,000 samples with 49 features representing students from various regions. The data was collected through a combination of automated tracking, periodic surveys, and manual input, focusing on three key categories: Sociodemographic Data,includes variables such as age, gender, socio-economic status, and educational background. Behavioral Data, captures metrics like login frequency, time spent on the platform, and course completion rates. Student Comments, analyzed through sentiment analysis to understand students' emotional responses. To ensure data quality, rigorous preprocessing was applied, including handling missing data through mean imputation and detecting outliers using Z-score analysis. The dataset was anonymized to protect student privacy, and the sampling strategy ensured a representative selection of students.

\subsection{ Exploratory data}

Exploratory data analysis (EDA) is a critical step before modeling and prediction, as it provides an initial overview of the key areas of interest within the dataset. This analysis helps identify patterns and relationships in the data, essential for accurate predictive analysis.Before we even start modeling and prediction, EDA is a critical step that provides an initial overview of the key areas of interest within the dataset. This analysis is not just a formality, but a necessity that helps us identify patterns and relationships in the data, which are essential for accurate predictive analysis. EDA involves the use of statistical techniques to understand the characteristics and interrelationships of the data, including univariate analysis, bivariate research, and multivariate analytics.
In sentiment analysis, EDA plays a crucial role by helping us comprehend text data's structure, patterns, and intricacies. This is vital for effective feature extraction and model selection. The EDA process typically begins with data inspection, followed by text length analysis, tokenization, stopword removal, word frequency analysis, n-gram analysis, and sentiment lexicon analysis, culminating in preliminary sentiment analysis.

\begin{figure}
\centering
\includegraphics[width=0.7\textwidth]{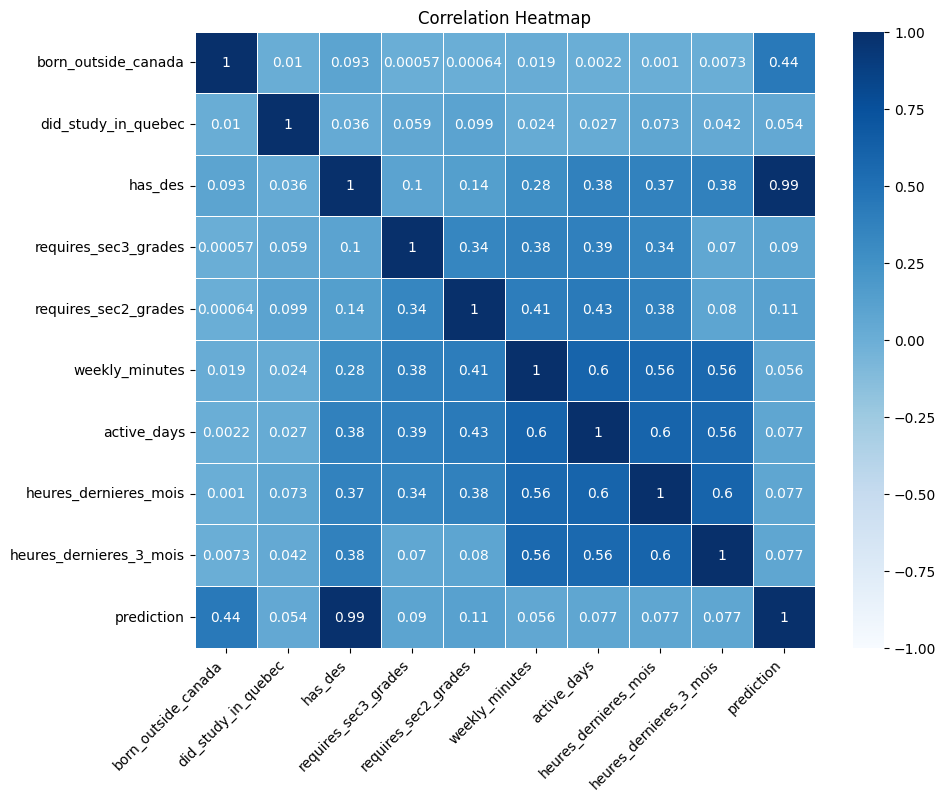}
\caption{Correlation Matrix for Identifying Attribute Interrelationships.} \label{correlation}
\end{figure}
The correlation heatmap in Figure~\ref{correlation} highlights the key relationships between sociodemographic factors, behavioral metrics, and early dropout risks in an online educational platforms. Notably, \textit{weekly\_minutes} and \textit{active\_days} show strong positive correlations with dropout prediction, emphasizing the critical role of student engagement in preventing dropout. Conversely, weaker correlations between certain sociodemographic variables and dropout risk suggest that these factors may influence dropout through indirect mechanisms or require further investigation. Significant interactions between \textit{weekly\_minutes} and \textit{active\_days} also suggest that increased engagement in both time and activity substantially reduces dropout likelihood. These findings indicate that predictive models can be enhanced by prioritizing strongly correlated features and that targeted interventions focusing on student engagement are likely to be most effective in preventing early dropout. The analysis highlights the importance of a comprehensive approach that combines both behavioral and sociodemographic data for accurate dropout prediction and prevention.

\subsection{Feature Engineering}
\begin{figure}[h]
    \centering    \includegraphics[width=0.7\textwidth]{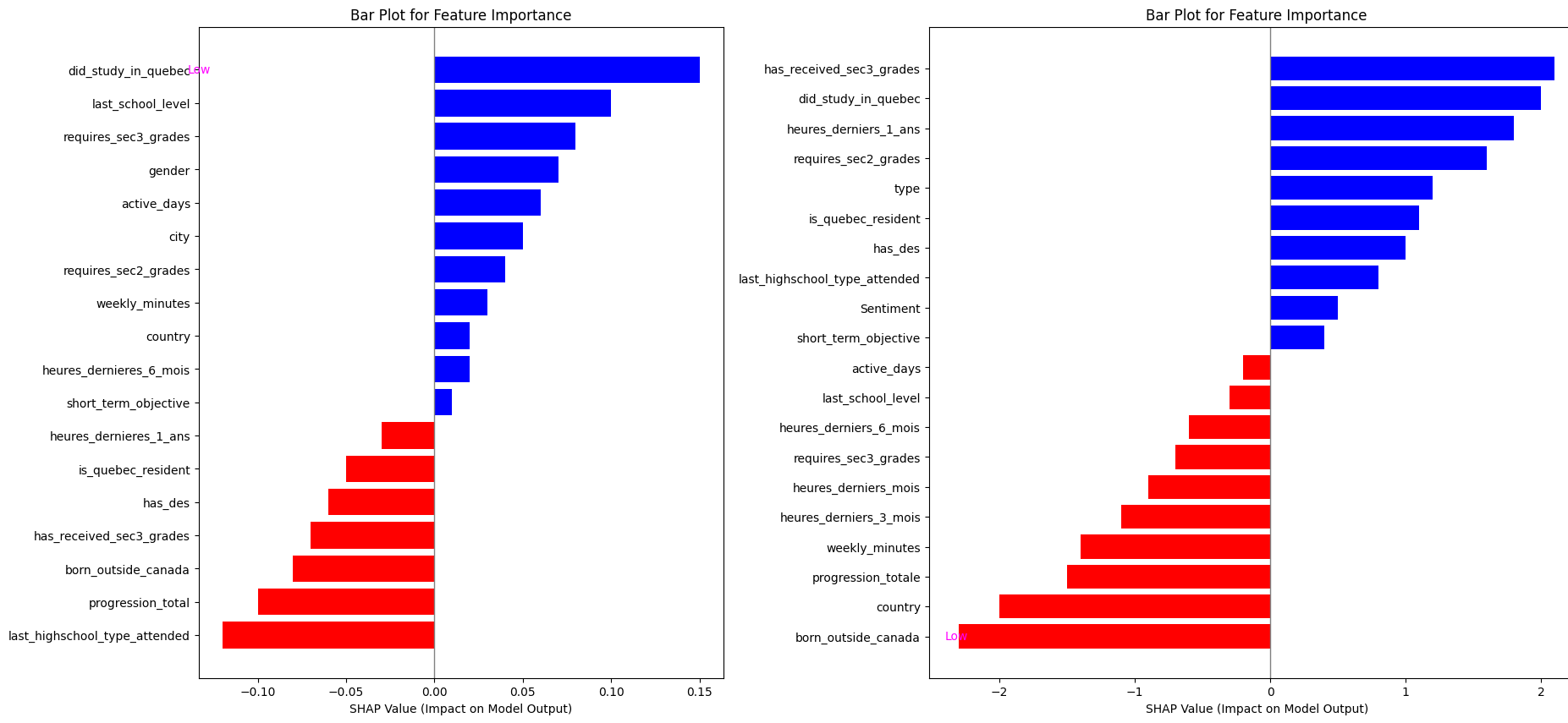}  
    \caption{Comparative Analysis of Feature Importance Across Two datasets.}
    \label{shap}
\end{figure}

The SHAP analysis of feature importance, depicted in Figure~\ref{shap}, provides a comprehensive overview of the critical features influencing early dropout in online education. This analysis hilghts the crucial role of feature engineering and selection in enhancing predictive models for student retention. The SHAP plots reveal the contribution of various features socio-demographic, behavioral, and sentiment related to the model's predictions, allowing us to assess not only the importance of individual features but also their influence on the model's output. This detailed study enables the development of more targeted and effective intervention strategies, which are essential for minimizing dropout rates.

In the dataset without sentiment (figure on the left), features such as \textit{progression\_totale} and \textit{heures\_derniers\_1\_ans} indicate high SHAP values. This indicates that student progress and recent activity hours are pivotal factors in predicting dropout risk. \textit{progression\_totale} reflects the overall advancement of the student in their coursework, suggesting that students who are not progressing as expected are at a higher risk of dropping out. Similarly, \textit{heures\_derniers\_1\_ans}, which captures the hours spent on learning activities in the last year, suggests that reduced engagement in recent times can be a strong predictor of dropout.

In the dataset with sentiment (figure on the right), additional features like \textit{Sentiment} are highlighted with significant SHAP values. The inclusion of sentiment data introduces a dimension of emotional and affective factors to the predictive model, revealing that students' emotional states, as inferred from their interactions, play a substantial role in determining dropout risk. For example, negative or neutral sentiment towards the learning material or platform may indicate disengagement or dissatisfaction, even if the student's progression appears satisfactory.

\subsection{Model training and hyper-parameters tuning}

 We apply various performance metrics displayed through confusion matrices, radar charts, and boxplots to demonstrate how different models and hyper-parameter tunings influence predictive accuracy.

The impact of hyperparameter tuning was evident in the significant improvement in model performance. For instance, the tuned XGBoost model achieved an accuracy of 84\%, outperforming the untuned model, which had an accuracy of 78\%. Additionally, precision and recall metrics also improved, demonstrating the importance of carefully selecting and optimizing hyperparameters in ML.

\begin{figure}[H]
    \centering
    \begin{minipage}{0.49\textwidth}
        \centering
        \includegraphics[width=\textwidth]{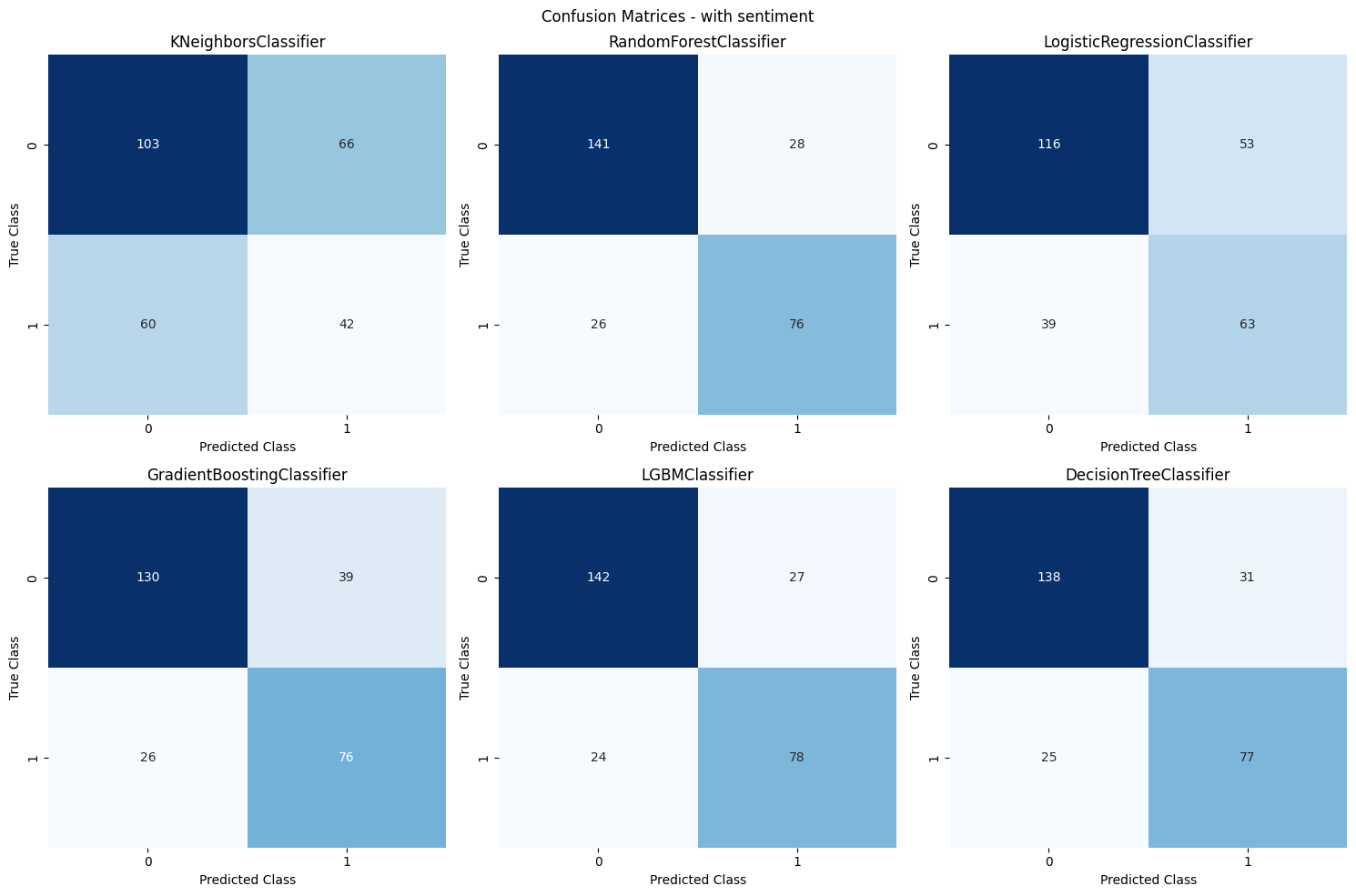}
        \caption{Comparison of Confusion Matrices for Predictive models(1).}
        \label{fig:confusion_1}
    \end{minipage}
    \hfill
    \begin{minipage}{0.49\textwidth}
        \centering
        \includegraphics[width=\textwidth]{confusion-new.png}
        \caption{Comparison of Confusion Matrices for Predictive models(2).}
        \label{fig:confusion_2}
    \end{minipage}
\end{figure}

Figures ~\ref{fig:confusion_1} and ~\ref{fig:confusion_2} present confusion matrices for various models under two different sets of hyper-parameters. These matrices allow us to assess each model's sensitivity and specificity in detecting dropout cases. A confusion matrix provides a breakdown of the model predictions into four outcomes—true positives (TP), true negatives (TN), false positives (FP), and false negatives (FN). Analyzing these outcomes helps in understanding the balance each model achieves between accurately identifying at-risk students (TP) and correctly confirming non-risk students (TN), while minimizing wrongful classifications (FP and FN).

In tuning hyper-parameters, we aime to optimize parameters such as the regularization strength in logistic regression, the depth of trees in decision tree-based models, and the kernel coefficients in SVMs. The effectiveness of these tunings is reflected in the shifts in TP and TN rates across the matrices, demonstrating how subtle changes in parameters can significantly affect model performance.

\subsection{Comparison of prediction performance} 
\begin{figure}[h]
    \centering
    \begin{minipage}{0.32\textwidth}
        \centering
        \includegraphics[width=\textwidth]{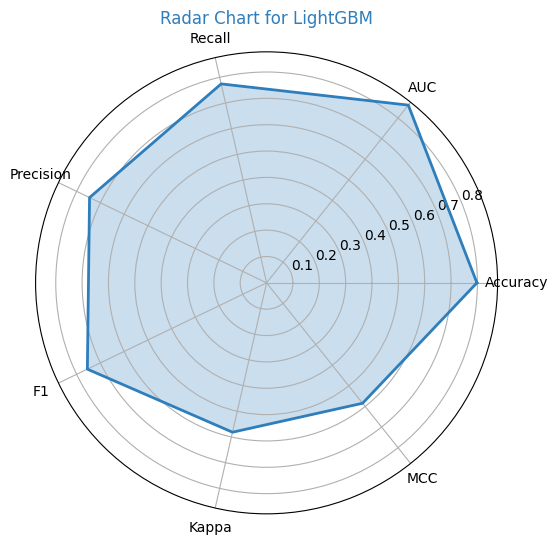}
        \caption{Radar Chart of Performance Metrics for LightGBM.}
        \label{radar}
    \end{minipage}
    \hfill
    \begin{minipage}{0.32\textwidth}
        \centering
        \includegraphics[width=\textwidth]{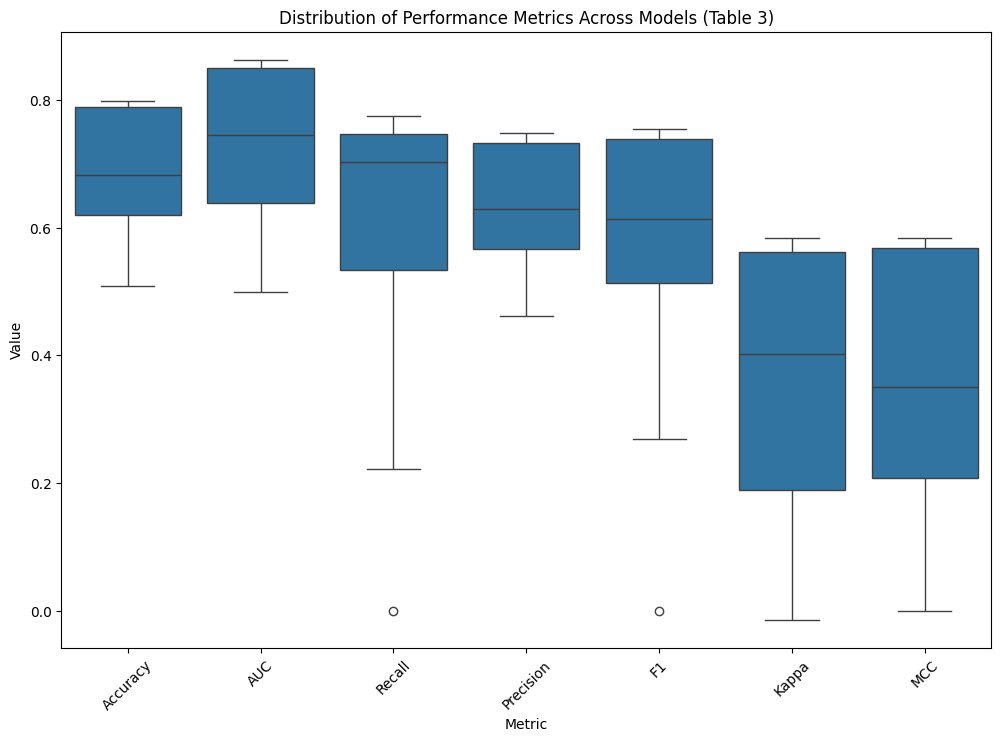}
        \caption{Boxplot Comparison of Performance Metrics Across Different Models(1).}
        \label{plot1}
    \end{minipage}
    \hfill
    \begin{minipage}{0.32\textwidth}
        \centering
        \includegraphics[width=\textwidth]{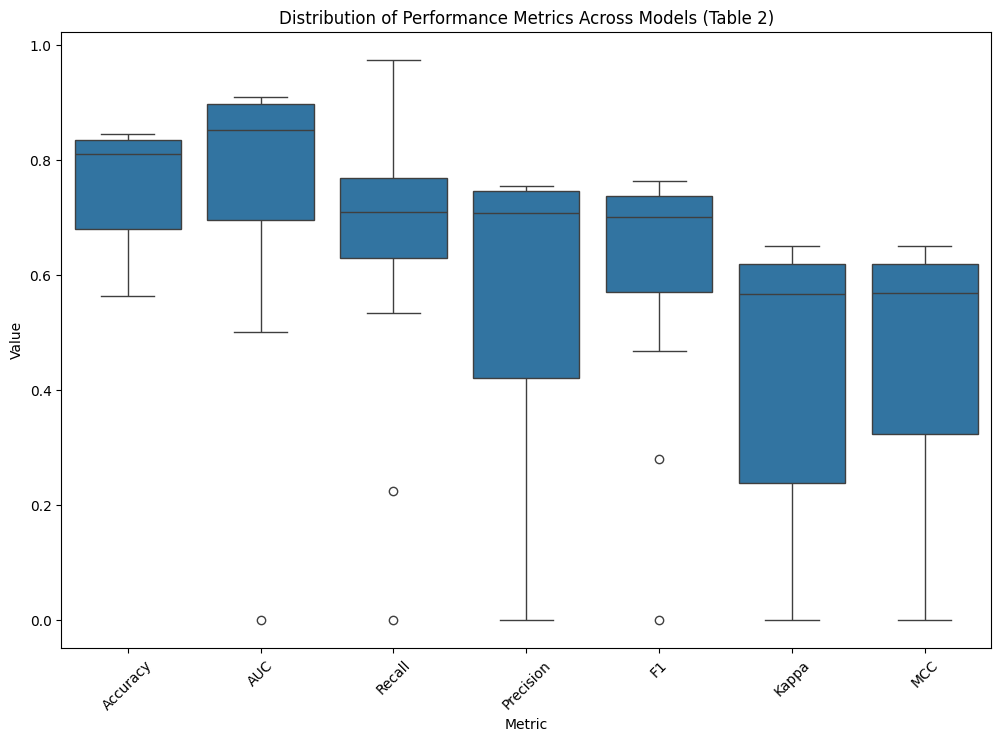}
        \caption{Boxplot Comparison of Performance Metrics Across Different Models(2).}
        \label{plot2}
    \end{minipage}
\end{figure}

Figure ~\ref{radar} uses a radar chart to display a comprehensive profile of LightGBM’s performance across multiple metrics, including Accuracy, Recall, Precision, F1 Score, AUC, MCC, and Kappa. This visualization aids in quickly assessing which aspects of predictive performance are strengths (e.g., high recall or precision) and which are weaknesses. A high recall indicates that LightGBM is effective in identifying most true dropout cases, but if precision is low alongside, the model may also be misclassifying many non-dropout students as at risk. The radar chart highlights that LightGBM, with its high recall (0.7) and decent AUC (0.7), is particularly well-suited for tasks where the primary goal is to identify as many at-risk students as possible. However, the lower precision (0.4) and moderate scores in F1 (0.5), MCC (0.4), and Kappa (0.3) suggest that while LightGBM is effective in identifying potential dropouts, it may also generate a relatively high number of false positives, which could lead to inefficient use of resources.

Figures 6 and 7 present boxplots illustrating the distribution of key performance metrics Accuracy, AUC, Recall, Precision, F1 Score, Kappa, and MCC across different ML models, providing insights into their variability, consistency, and overall suitability for predicting school dropout. The boxplots show that Accuracy and AUC have relatively high median values, around 0.8 to 0.9, with tight interquartile ranges, indicating that most models effectively classify students and distinguish between those likely to drop out and those who are not. However, Recall shows greater variability, with medians around 0.6 to 0.7 and outliers below 0.4, suggesting inconsistencies in detecting all at-risk students, which could impact dropout prevention strategies. Precision shows median values around 0.6 but with a wider interquartile range, highlighting variability in correctly identifying actual dropouts versus false positives. The F1 Score, also around 0.6, reflects the balance between Recall and Precision, emphasizing the need for models that can consistently manage this trade-off. Kappa and MCC, with lower medians around 0.4 to 0.6, highlight opportunities for improvement in ensuring predictive reliability.  Integrating sentiment analysis could help address these inconsistencies, particularly in Recall and Precision, leading to more reliable and effective dropout prevention strategies.

\section{Ablation Study}

This ablation study highlights the impact of sentiment analysis on our predictive models, focusing on its practical implications when processing socio-demographic and behavioral data. By comparing model performances with and without sentiment data, we aim to determine how the textual analysis of student comments affects predictive accuracy. Specifically, we evaluate metrics such as accuracy, precision, recall, and F1-score to assess the contribution of sentiment features. This approach helps identify the distinct benefits of incorporating sentiment analysis, enhancing the reliability and precision of dropout predictions.

\begin{itemize}
    \item \textbf{Without Sentiment Analysis:} This module evaluates the model's performance using only socio-demographic and behavioral data. This serves as a baseline to understand the model's capabilities in the absence of textual analysis \cite{Zerkouk2024AML}.
    \item \textbf{With Sentiment Analysis:} In this module, we integrate sentiment analysis data to assess how this addition enhances the model's ability to identify sentiment-related dropout risks. The inclusion of sentiment analysis significantly improves the performance of models like Naive Bayes and SVM due to their proficiency in processing and interpreting textual and experiential data from student comments \cite{{Sahoo2023}}.
\end{itemize}

Effects of sentiment analysis in predicting student dropouts:
\begin{itemize}
    \item \textbf{Temporal Dynamics:} We examine how the timing of sentiment expression impacts model predictions, analyzing whether sentiments expressed early in the academic term are more predictive of dropout than those expressed later. Our analysis revealed that sentiments expressed during the early weeks of the academic term have a disproportionately large impact on the likelihood of dropout. For instance, students who expressed negative sentiments within the first month of the course were found to have a significantly higher dropout rate than those who expressed similar sentiments later in the term. This suggests that early detection and intervention are crucial for preventing dropout
    \item \textbf{Interaction Dynamics:} We investigate how sentiment data interacts with other predictors and identify specific instances where the combination of sentiment and other variables significantly contributes to predictive accuracy. Our study revealed that the interaction between student engagement and sentiment plays a pivotal role in determining dropout risk. Specifically, students with low engagement and negative sentiment were at the highest risk of dropping out.
\end{itemize}

\begin{figure}[ht]
    \centering    \includegraphics[width=0.7\textwidth]{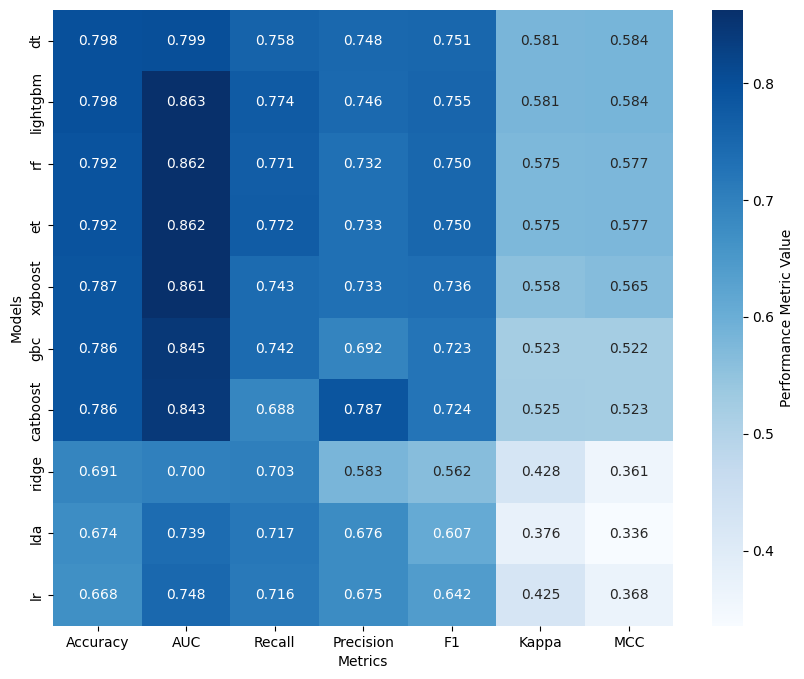}
    \caption{Heatmap of Performance Metrics Across Different Models performances.}
    \label{fig1}
\end{figure}

\subsection*{Discussion}
As shown in Figure~\ref{fig1}, significant performance discrepancies are evident when sentiment analysis is excluded, particularly with models like Naive Bayes and SVM. For instance, the exclusion of sentiment analysis results in notable decreases in SVM’s accuracy (by 0.65\%) and F1 score (by 0.50). For instance, the inclusion of sentiment analysis leads to substantial improvements in several key metrics for these models. Naive Bayes and SVM show enhancements in recall (0.75 and 0.56, respectively) and accuracy (0.85 and 0.52, respectively). This demonstrates the critical role of sentiment analysis in enhancing models’ capabilities to accurately predict dropout risks based on student comments.

In contrast, models like XGBoost show minimal performance changes, suggesting their predictive capabilities are less dependent on sentiment analysis. This variability underscores the importance of selecting appropriate features based on the specific model used and highlights how sentiment analysis can significantly enhance the performance of certain models, proving a valuable addition to improving predictive accuracy and identifying at-risk students.

This ablation study not only highlights the differential impact of sentiment analysis on various models but also provides crucial insights into how each component contributes to overall predictive performance. By identifying these factors, we can adjust our predictive models to capture the most informative features, thereby significantly enhancing their effectiveness in the field of education.

Early detection of negative sentiments allows for rapid interventions that can prevent student dropout while promoting perseverance by providing the necessary and personalized interventions to overcome challenges, ultimately leading to better academic outcomes.

\section{Conclusion}
Student dropout is one of the most significant challenges faced by distance learning platforms. In this paper, we developed a model that demonstrates the potential of integrating sentiment analysis into dropout prediction models for these platforms. Our proposed model combines sentiment analysis data, extracted using the BERT model, with sociodemographic and behavioral data to form a comprehensive model for predicting student dropout. By incorporating insights from sentiment analysis, this approach addresses limitations of previous research related to the omission of emotional and experiential factors.

The robustness of our prediction model, which achieved an 84\% accuracy using the XGBoost algorithm, underscores the importance of feature richness. This highlights the potential benefits of our approach in shaping targeted interventions, support strategies, and encouraging student perseverance. Our research contributes to a more nuanced understanding of the factors affecting student retention on distance learning platforms by integrating sentiment analysis with other data sources and demonstrating its potential to enhance existing prediction models. Future research could explore the application of this framework in different educational contexts. 

%
%
%

%
%
%

\begin{thebibliography}{8}

\bibitem{LeDevoir2023}
Le Devoir: Plus de 10 000 décrocheurs scolaires au Québec. In: Le Devoir, \url{https://www.ledevoir.com/opinion/idees/753858/milieux-defavorises-plus-de-10-000-decrocheurs-scolaires-au-quebec} (2023).

\bibitem{Berens2018}
Berens, J., et al.: Early Detection of Students at Risk – Predicting Student Dropouts Using Administrative Student Data and Machine Learning Methods. In: SSRN Electronic Journal (2018), n. pag.

\bibitem{Pek2023}
Pek, R. Z., et al.: The Role of Machine Learning in Identifying Students At-Risk and Minimizing Failure. In: IEEE Access, vol. 11, pp. 1224-1243 (2023).

\bibitem{Niyogisubizo2022}
Niyogisubizo, J., et al.: Predicting student's dropout in university classes using a two-layer ensemble machine learning approach: A novel stacked generalization. In: Computers in Education: Artificial Intelligence, vol. 3, 100066 (2022).

\bibitem{Karabacak2023}
Karabacak, E. S., Yaslan, Y.: Comparison of Machine Learning Methods for Early Detection of Student Dropouts. In: 2023 8th International Conference on Computer Science and Engineering (UBMK), pp. 376-381 (2023).

\bibitem{RodaSegarra2024}
Roda-Segarra, J., et al.: Effectiveness of Artificial Intelligence Models for Predicting School Dropout: A Meta-Analysis. In: Multidisciplinary Journal of Educational Research (2024), n. pag.

\bibitem{Adnan2021}
Adnan, M., Habib, A., Ashraf, J., Mussadiq, S., Raza, A. A., Abid, M., et al.: Predicting at-risk students at different percentages of course length for early intervention using machine learning models. In: IEEE Access, vol. 9, pp. 7519-7539 (2021). \doi{10.1109/ACCESS.2021.3049446}

\bibitem{Sani2020}
Sani, N. S., Nafuri, A. F. M., Othman, Z. A., Nazri, M. Z. A., Mohamad, K. N.: Dropout prediction in higher education among B40 students. In: International Journal of Advanced Computer Science and Applications, vol. 11, no. 11 (2020). \doi{10.14569/IJACSA.2020.0111169}


\bibitem{Kotsiantis2003}
Kotsiantis, S. B., Pierrakeas, C. J., Pintelas, P. E.: Preventing Student Dropout in Distance Learning Using Machine Learning Techniques. In: International Conference on Knowledge-Based Intelligent Information and Engineering Systems (2003). \url{https://api.semanticscholar.org/CorpusID:6281847}

\bibitem{Kemper2020}
Kemper, L., Vorhoff, G., Wigger, B. U.: Predicting student dropout: A machine learning approach. In: European Journal of Higher Education, vol. 10, pp. 28–47 (2020).


\bibitem{Solis2018}
Solís, M., Moreira, T. M. B., Gonzalez, R., Fernandez, T., Hernandez, M.: Perspectives to predict dropout in university students with machine learning. In: 2018 IEEE International Work Conference on Bioinspired Intelligence (IWOBI), pp. 1

\bibitem{Karabacak2024}
Pandian, B. R., Abdul Aziz, A., Subramaniam, H., Sutan Ahmad Nawi, H.: Exploring the Role of Machine Learning in Forecasting Student Performance in Education: An In-Depth Review of Literature. In: Multidisciplinary Reviews, pp. 376-381 (2024).


\bibitem{Khanam2023}
Khanam, Z.: Sentiment Analysis of User Reviews in an Online Learning Environment: Analyzing the Methods and Future Prospects. In: European Journal of Education and Pedagogy (2023). \url{https://api.semanticscholar.org/CorpusID:258121473}

\bibitem{Baragash2021}
Baragash, R. S., Aldowah, H.: Sentiment Analysis in Higher Education: A Systematic Mapping Review. In: Journal of Physics: Conference Series, vol. 1860 (2021). \url{https://api.semanticscholar.org/CorpusID:233849186}

\bibitem{Zerkouk2024AML}
Zerkouk, M., Mihoubi, M., Chikhaoui, B., Wang, S.: A machine learning based model for student’s dropout prediction in online training. In: \textit{Education and Information Technologies}, 2024. \url{https://api.semanticscholar.org/CorpusID:267478838}

\bibitem{Sahoo2023}
Sahoo, C., Wankhade, M., Singh, B. K.: Sentiment Analysis Using Deep Learning Techniques: A Comprehensive Review. In: International Journal of Multimedia Information Retrieval, vol. 12, pp. 1-23 (2023). \url{https://api.semanticscholar.org/CorpusID:265353039}





\end{thebibliography}
%

\end{document}